\documentclass[a4paper,fleqn]{cas-sc}

\usepackage[authoryear,longnamesfirst]{natbib}

\makeatletter
\let\cas@beginabstract\abstract      % \abstract is the *begin* macro
\let\cas@endabstract  \endabstract   % \endabstract is the *end*  macro
\makeatother
\usepackage{arabtex}
\usepackage{utf8}
\setcode{utf8}

\makeatletter
\let\abstract   \cas@beginabstract   % restore begin
\let\endabstract\cas@endabstract     % restore end
\makeatother
\usepackage{graphicx}
\usepackage{tabularx}
\usepackage{booktabs}
\usepackage{amssymb}
\usepackage{pifont}
\usepackage{tikz}
\usetikzlibrary{arrows.meta, positioning}

\newcommand{\xmarkred}{\textcolor{red}{\ding{55}}}

\def\tsc#1{\csdef{#1}{\textsc{\lowercase{#1}}\xspace}}
\tsc{WGM}
\tsc{QE}
\tsc{EP}
\tsc{PMS}
\tsc{BEC}
\tsc{DE}

\shorttitle{Advancing Arabic Reverse Dictionary Systems}
\shortauthors{Sibaee et~al.}

\title [mode=title]{Advancing Arabic Reverse Dictionary Systems: A Transformer-Based Approach with Dataset Construction Guidelines}
\author[1]{Serry Sibaee}
\cormark[1]
\fnmark[1]
\ead{ssibaee@psu.edu.sa}

\author[3]{Samar Ahmed}
\ead{Samar.sass6@gmail.com}

\author[2]{Abdullah Al Harbi}
\ead{aamalharbe@kau.edu.sa}

\author[1]{Omer Nacar}
\ead{onajar@psu.edu.sa}

\author[1]{Adel Ammar}
\ead{aammar@psu.edu.sa}

\author[1]{Yasser Habashi}
\ead{yalhabashi@psu.edu.sa}

\author[1]{Wadii Boulila}
\fnmark[2]
\ead{wboulila@psu.edu.sa}

\affiliation[1]{organization={College of Computer \& Information Sciences, Prince Sultan University},
                city={Riyadh},
                country={Saudi Arabia}}%
                
\affiliation[2]{organization={Faculty of Computing and Information Technology, King Abdulaziz University},
                city={Jeddah},
                country={Saudi Arabia}}%
                
\affiliation[3]{organization={Independent Researcher},                city={Riyadh},
country={Saudi Arabia}}%

\begin{document}
\begin{abstract}
This study addresses the critical gap in Arabic natural language processing by developing an effective Arabic Reverse Dictionary (RD) system that enables users to find words based on their descriptions or meanings. We present a novel transformer-based approach with a semi-encoder neural network architecture featuring geometrically decreasing layers that achieves state-of-the-art results for Arabic RD tasks. Our methodology incorporates a comprehensive dataset construction process and establishes formal quality standards for Arabic lexicographic definitions. Experiments with various pre-trained models demonstrate that Arabic-specific models significantly outperform general multilingual embeddings, with ARBERTv2 achieving the best ranking score (0.0644). Additionally, we provide a formal abstraction of the reverse dictionary task that enhances theoretical understanding and develop a modular, extensible Python library (RDTL) with configurable training pipelines. Our analysis of dataset quality reveals important insights for improving Arabic definition construction, leading to eight specific standards for building high-quality reverse dictionary resources. This work contributes significantly to Arabic computational linguistics and provides valuable tools for language learning, academic writing, and professional communication in Arabic.
\end{abstract}
\begin{graphicalabstract}
\includegraphics[width=0.75\textwidth]{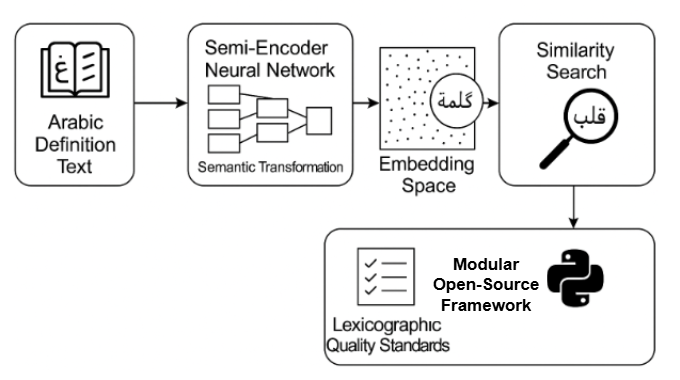}
\end{graphicalabstract}
\begin{highlights}
\item We propose a semi-encoder neural network with geometrically decreasing hidden layers, achieving state-of-the-art performance on Arabic reverse dictionary tasks through efficient semantic representation.
\item We establish formal quality standards for Arabic lexicographic definitions, providing a systematic framework to enhance dataset construction for reverse dictionary applications.
\item We present a formal abstraction of the reverse dictionary task, offering a rigorous theoretical framework for embedding-based word retrieval systems and facilitating reproducibility and future research.
\item We develop RDTL, a modular and extensible Python library with configurable training pipelines for reverse dictionary tasks, released as open-source to support the research community.\footnote{\url{https://github.com/serrysibaee/reverse_dictionary/tree/main}}
\end{highlights}
\maketitle

% Main body
% \input{script/name}
\section{Introduction}

Recent achievements in natural language processing (NLP) have played an essential role in advancing various semantic understanding tasks. Of these, the Reverse Dictionary (RD) is a prominent semantic task that tackles the problem of finding a word from a given description or meaning.  Whereas a traditional forward dictionary associates a given word with its formal definition, RD solves the inverse problem of assisting users identify a word when given its meaning. This function has many practical applications in different fields and helps us tackle common linguistic and cognitive issues.

Perhaps the most famous use of RD is mitigating the phenomenon known as the "tip of the tongue" (TOT), the frustrating stage where one remembers a meaning or some synonyms, yet can not reach the specific word itself  \cite{brown1966tip}. Using RD to search by meaning of word enables users to create much more accurate language. In academic writing, for example, RD can assist a researcher in choosing the most precise term to convey a complicated thought \cite{pilehvar-2019-importance}. In contract drafting and other specialized domains, RD can assist professionals in selecting precise terminology by mapping natural language descriptions to semantically appropriate words, ensuring clarity and accuracy in legal and business communication \cite{Siddique2022BuildingAR}. In addition, RD also helps learn and teach languages. For those studying a second language, such systems can serve as an efficient means of strengthening their vocabulary by assisting them in finding new words and learning their subtle meanings  \cite{mane2022wordalchemytransformerbasedreversedictionary}. This is what makes RD an integral tool in educational institutions, and the creation of language-learning apps.

While RDs have been developed for languages such as English, French, or Chinese \cite{yan-etal-2020-bert}, work to develop them for Arabic has been slower \cite{mickus-etal-2022-semeval}. This is due to the complexity of Arabic morphology, diglossia (the coexistence of formal and colloquial dialects), and the frequent omission of diacritics, which can introduce ambiguity. For instance, KSAA-RD shared tasks in 2023 and 2024 focus on this issue and seek to encourage the generation of Arabic RD. But challenges persist, given the intricacies of Arabic morphology and the limited availability of linguistic resources. Hence, there is a strong potential for further development of sound Arabic RD that will contribute to Arabic natural language processing and allow users to benefit from the same lexical precision available in other major languages.

This work aims to advance the development of reliable RD models for Arabic by introducing several key contributions:

\begin{itemize}
    \item We propose a semi-encoder neural network with geometrically decreasing hidden layers, achieving state-of-the-art performance on Arabic reverse dictionary tasks through efficient semantic representation.
    
    \item We establish formal quality standards for Arabic lexicographic definitions, providing a systematic framework to enhance dataset construction for reverse dictionary applications.
    
    \item We present a formal abstraction of the reverse dictionary task, offering a rigorous theoretical framework for embedding-based word retrieval systems and facilitating reproducibility and future research.
    
    \item We develop RDTL, a modular and extensible Python library with configurable training pipelines for reverse dictionary tasks, released as open-source to support the research community.\footnote{\url{https://github.com/serrysibaee/reverse_dictionary}}
\end{itemize}

In this paper, we detail our approach to constructing an Arabic RD model, beginning with an elaboration of the challenges and the need to propose a new solution in Section 2. This is followed by a description of our methodology in Section 3. Sections 4,5 presents a thorough evaluation of the model's performance on various tasks, analyzing the results and highlighting key findings in the models and dataset. Finally, Section 6 concludes the paper, discussing the implications of our work and outlining promising directions for future research in this rapidly evolving field.

\section{Related Work}

In this section, we provide an overview of existing reverse dictionary research. We begin by discussing traditional approaches based on semantic resources such as WordNet. Next, we describe neural network methods that rely on recurrent architectures. Finally, We focus on transformer-based approaches and their advancements in this domain.

\subsection{Traditional (WordNet-based) Approaches}
Early reverse dictionary systems often relied on semantic analysis performed on input phrases, leveraging resources like WordNet to identify and compare word senses. In these systems, words and phrases were typically mapped onto a semantic space, and distance-based similarity measures were used to find the best-matching candidates.  \cite{pilehvar-2019-importance} perform semantic analysis on input phrases using semantic similarity measures to represent words as vectors in a semantic space. The semantic space is created with the assistance of WordNet. The system then applies algebraic analysis to select a sample of candidate words. These candidate words undergo a filtering process and a ranking phase before being presented. It uses a similarity measure between a word and an input phrase based on a distance-based similarity metric.  \cite{thorat-choudhari-2016-implementing} employ the idea that the significance of a word's meaning to a definition is proportional to its frequency across various definitions. The meaning is extracted from the content words within the phrase. The input phrase is split into its component words, and a graph-based search is implemented through related words. A distance-based similarity measure computes the words that best represent the meaning of the input phrase. A graph encodes the relationships between words in its connectivity matrix, on which the similarity measures are computed.

\subsection{Traditional Neural Network Methods}
In contrast, several studies emphasize the potential of neural networks \cite{fatima2022natural}, leveraging their advanced modeling techniques to capture the context and semantic meaning of text more effectively. \cite{pilehvar-2019-importance} present experiments using a neural network-based reverse dictionary system that focuses on distinguishing between the different meanings of a word to achieve more accurate semantic understanding. To address this issue, two main neural architectures are employed for processing definitions: the Bag-of-Words (BoW) model and the Recurrent Neural Network (RNN) model, aiming to resolve the challenges associated with polysemy. \cite{bendahman-etal-2022-bl} use sequential models that integrate various neural networks, starting with embedding layers and progressing through dense layers, Bidirectional Long Short-Term Memory (BiLSTM) networks, and LSTM networks. All glosses are pre-processed to ensure optimal representation of word meanings. The three embedding techniques used are “char,” “sgns,” and “electra.” The results show that models trained with character-based or contextualized embeddings outperform those using Skip-Gram word embeddings. Their model effectively learns to map arbitrary-length phrases to fixed-length continuous-valued word vectors.  \cite{chen-zhao-2022-unified} present a model that functions as a neural dictionary with two-way indexing and querying, embedding both words and definitions within a shared semantic space. The model focuses on two primary tasks: retrieving words based on their definitions and generating definitions for given words. It uses separate encoder and decoder networks for words and definitions, complemented by a shared layer that aligns them within the same representation space. The model demonstrates impressive performance on established benchmarks without requiring additional resources, and human evaluations indicate a preference for its outputs. \cite{agrawal2021reverse} combine the Continuous Bag-of-Words (CBOW) model, which considers a fixed number of surrounding words without accounting for their order, with a Recurrent Neural Network (RNN), which uses only the preceding context while preserving word order, to create a reverse dictionary that effectively captures both context and word order.

 \cite{malekzadeh2021predictpersianreversedictionary} develop four different architectures for a Persian reverse dictionary using (phrase, word) pairs sourced from online Persian dictionaries. These architectures include a Bag-of-Words (BoW) model, RNN, LSTM, and BiLSTM models with additive attention. Each model maps descriptive phrases to their corresponding words, with the additive attention mechanism notably enhancing their ability to generate relevant word suggestions. The method presented in \cite{zhang2019multichannelreversedictionarymodel} employs a multi-channel reverse dictionary model based on a Bidirectional LSTM with an attention mechanism to encode input queries. Multiple predictors, such as a POS tag predictor and a morpheme predictor, are incorporated to enhance the accuracy of identifying target words from given definitions. Similarly, the work in \cite{qi-etal-2020-wantwords} adopts a comparable approach, introducing an innovative, open-source, online reverse dictionary system. This contribution marks a significant advancement in the field.

\subsection{Transformer-based Approaches}
More recent research utilizes transformer architectures capable of processing longer sequences and richer context. By incorporating specialized tokenization strategies and attention mechanisms. These methods significantly improve semantic representation and retrieval accuracy in reverse dictionary applications \cite{alshattnawi2024beyond}. \cite{hedderich2019usingmultisensevectorembeddings} introduce attention mechanisms to integrate multi-sense embeddings, which are vector representations capturing different meanings (senses) of a word using word2vec and GloVe \cite{roman2021citation}. These embeddings are combined with LSTM models and contextual word embeddings, such as Bidirectional Encoder Representations from Transformers (BERT), to enhance performance in the reverse dictionary task. The findings demonstrate significant improvements in both input sequences and target representations, alongside valuable insights into sense distributions and learned attention patterns.
The KSAA-RD shared task \cite{al-matham-etal-2023-ksaa} focuses on developing a Reverse Dictionary (RD) system specifically for Arabic with cross-lingual reverse dictionaries (CLRD), aiming to advance Arabic natural language processing, foster research, and create tools to enhance understanding and usage of the Arabic language. The winning teams are, respectively, modeled by \cite{Elbakry_2023}. They leverage multiple BERT-based pre-trained models for RD with output averaging, including camelBERT-MSA, camelBERT-Mix, MARBERTv2, and AraBERTv2. They achieve a ranking of 24.20 using ELECTRA embeddings when ensemble embeddings from camelBERT-MSA and MARBERTv2 models are employed. For the CLRD task, the same models attain a rank of 12.70 with ELECTRA embeddings. \cite{sibaee-etal-2023-qamosy} implemented a two-stage approach. In the first stage, the Sentence Transformer (SBERT) is used to generate fixed-length embeddings. A Semi-Decoder model is employed in the second stage to transform the inputs into output vectors with two dimensions: ELECTRA (265d) and SGNS (300d). The model achieves high RD rankings, with the best result obtained using ELECTRA embeddings, reaching a rank of 28.10. \cite{qaddoumi-2023-abed} introduced a modified BERT Multilingual model for both RD and CLRD tasks with data augmentation due to the limited size of the available data. Data augmentation techniques include synonym replacement, random word insertion, deletion, swapping within sentences in English, and random word deletion and swapping in Arabic. They achieve a rank of 28.50 for the RD task and 28.10 for the CLRD task with ELECTRA embeddings. Finally,  \cite{taylor-2023-uwb} employ a rule-based approach for RD and CLRD tasks. They do not use neural networks but use SGNS and ELECTRA models for semantic vector representations. They build a dataset-based dictionary and expand it using glosses, focusing on normalization, verb inflection, and gloss adjustments while omitting certain parts of speech. The dictionary-based approach with SGNS embeddings achieves a score of 43.8 for the RD task, lower than the baseline model, and 48.87 for the CLRD task.

\cite{alshammari-etal-2024-ksaa}, in the KSAA-RD shared task, integrated Word Sense Disambiguation (WSD) specifically for Contemporary Arabic into Reverse Dictionary (RD) tasks, enhancing the clarification of word meanings within context.  \cite{sibaee-etal-2024-asos-ksaa} used a semi-encoder structure for reverse dictionary tasks, while a two-stage neural network was employed for word sense disambiguation. In the first stage, each input was processed independently through a neural network, and the resulting outputs were then concatenated and passed through another similar network. Various sentence transformer models, including distiluse-base-multilingual-cased-v1, MiniLM-L12-v2, and mpnet-base-v2, were evaluated for their suitability in text encoding tasks alongside the AraBERTv2 model. Using ELECTRA with this model, their system achieved the best rank and an MSE score of 0.0644 and 0.059. \cite{chen-etal-2024-cher} developed a multi-task framework combining RD, definition generation, and reconstruction using transformers. They explored three segmentation strategies: the whitespace tokenizer (as in the original work), the AraBERTv2 tokenizer based on Farasa, and the CAMeLBERT tokenizer, to segment the definitions and pair them with the provided embeddings, regardless of the model from which they were derived. The AraBERT (Farasa) tokenizer yielded promising performance across metrics in the development phase, with a ranking score of 0.4834. \cite{alheraki-meshoul-2024-baleegh} leveraged AraT5 V2 with SentencePiece tokenization, combining glosses and Wikipedia examples. A word and its definition were combined into a single input string, embedded using the multilingual-22-125 model, and cosine similarity was used for vector search to retrieve contextually relevant examples. They trained two models: the first was on glosses only, while the second was on glosses merged with the retrieved examples. Their methodology achieved the highest-ranking score of 0.1781 using ELECTRA embeddings with glosses only.  \cite{alharbi-2024-mission}, like  \cite{alheraki-meshoul-2024-baleegh}, utilized the AraT5 V2 model for the Arabic RD task with gloss-to-embedding mapping, employing three different architectures of contextualized word embeddings: AraELECTRA, AraBERTv2, and camelBERT-MSA. This approach achieved a ranking score of 0.2482 by fine-tuning gloss-based embeddings with the AraT5 V2 model.

Table~\ref{tab:related_works_comp} provided a comprehensive comparison of the various reverse dictionary approaches, highlighting their model types, language focus, and key features.

\begin{table*}[h]
\centering
\resizebox{\textwidth}{!}{
\begin{tabular}{c c c c c c}
\hline
\textbf{Work} & \textbf{Model Type} & \textbf{Primary Language Focus} & \textbf{Attention Mechanism} & \textbf{Multi-Channel} & \textbf{Pre-trained Base} \\ 
\hline
WordNet-based \cite{mendez2013reverse} & Traditional & English & \xmarkred & \xmarkred & \xmarkred \\
Graph-based \cite{thorat-choudhari-2016-implementing} & Traditional & English & \xmarkred & \xmarkred & \xmarkred \\
BoW/RNN Hybrid \cite{pilehvar-2019-importance} & Neural Network & English & \xmarkred & \xmarkred & \xmarkred \\
BiLSTM \cite{bendahman-etal-2022-bl} & Neural Network & English & \xmarkred & \xmarkred & Character/ELECTRA/SGNS \\
Unified Model \cite{chen-zhao-2022-unified} & Neural Network & English & \xmarkred & \checkmark & \xmarkred \\
CBOW-RNN \cite{agrawal2021reverse} & Neural Network & English & \xmarkred & \checkmark & \xmarkred \\
BiLSTM+Attention \cite{malekzadeh2021predictpersianreversedictionary} & Neural Network & Persian & \checkmark & \xmarkred & \xmarkred \\
Multi-channel BiLSTM \cite{qi-etal-2020-wantwords} & Neural Network & Multi-lingual & \checkmark & \checkmark & \xmarkred \\
Multi-sense with BERT \cite{hedderich2019usingmultisensevectorembeddings} & Transformer & English & \checkmark & \checkmark & BERT \\
Ensemble BERT \cite{Elbakry_2023} & Transformer & Arabic & \checkmark & \checkmark & camelBERT/MARBERT/AraBERT \\
SBERT+Semi-Decoder \cite{sibaee-etal-2023-qamosy} & Transformer & Arabic & \checkmark & \checkmark & SBERT/ELECTRA/SGNS \\
BERT Multilingual \cite{qaddoumi-2023-abed} & Transformer & Arabic/English & \checkmark & \xmarkred & BERT Multilingual \\
Rule-based Vector \cite{taylor-2023-uwb} & Traditional & Arabic & \xmarkred & \xmarkred & SGNS/ELECTRA \\
Word Sense Integration \cite{alshammari-etal-2024-ksaa} & Transformer & Arabic & \checkmark & \xmarkred & \xmarkred \\
Semi-encoder \cite{sibaee-etal-2024-asos-ksaa} & Transformer & Arabic & \checkmark & \checkmark & distiluse/MiniLM/mpnet/AraBERT \\
Multi-task Framework \cite{chen-etal-2024-cher} & Transformer & Arabic & \checkmark & \checkmark & AraBERT/CAMeLBERT \\
AraT5 V2 \cite{alheraki-meshoul-2024-baleegh} & Transformer & Arabic & \checkmark & \checkmark & multilingual-22-125 \\
AraT5 V2 Fine-tuned \cite{alharbi-2024-mission} & Transformer & Arabic & \checkmark & \checkmark & AraELECTRA/AraBERT/camelBERT \\
\hline
\end{tabular}
}
\caption{Comparison of Reverse Dictionary approaches by model type, language focus, and features.}
\label{tab:related_works_comp}
\end{table*}

 As shown in Table~\ref{tab:related_works_comp}, there has been a clear evolution from traditional methods to neural networks and finally to transformer-based architectures, with increasing adoption of attention mechanisms and multi-channel approaches, particularly for Arabic language models.

\section{Methodology}
This section describes the methodology employed in our experiments, starting with the dataset and its augmentation, followed by the model architecture and training approach.

\subsection{Data Description}

The dataset consists of 31,372 training samples, 3,921 validation samples, and 3,922 test samples. Each sample is represented as a pair \( (\text{def}, \text{word}) \), where both the definition and target word are provided in text and embedding formats. For this task, we were given the Electra embeddings (256-dimensional) \( e: \text{Word} \to \mathbb{R}^{256} \) \cite{antoun-etal-2021-araelectra}. 

To enhance model performance, the dataset was expanded by incorporating approximately 84,000 additional samples containing Electra embeddings from an external source. Following prior analysis, the training and development sets were merged, and no separate results are reported for the development subset.

Table \ref{tab:dataset_samples} depicts a sample from the dataset.
\begin{table*}[h]
\centering
\small
\renewcommand{\arraystretch}{1.1}
\begin{tabular}{|p{3.2cm}|p{6.2cm}|>{\raggedright\arraybackslash}p{4.1cm}|}
\hline
\textbf{Term} & \textbf{Definition} & \textbf{Translation} \\
\hline
\<هَجْر> & \<القَطيعة، وهو ضدّ الوَصْل> & Estrangement; it is the opposite of union. \\
\hline
\<جماعة الدِّيوان> & \<تطلق على النقاد الثلاثة الذين كوّنوا مدرسة الديوان> & A group of three critics who formed the Diwan School: Al-Aqqad, Al-Mazini, and Shukri. They agreed on poetic vision and artistic foundations. \\
\hline
\<جلَخ السَّيلُ الواديَ> & \<جرف جوانبَه وملأه ماء> & The flood eroded its sides and filled it with water. \\
\hline
\end{tabular}
\caption{Sample entries from the reverse dictionary dataset.}
\label{tab:dataset_samples}
\end{table*}

\subsection{The proposed Neural Network Architecture}

Let \( D \) be a dataset consisting of pairs \( (\text{def}_i, \text{word}_i) \), where each definition \(\text{def}_i\) is associated with a corresponding word \(\text{word}_i\). We define two embedding functions:
\begin{itemize}
    \item \( f: \text{Def} \to \mathbb{R}^d \) maps a definition to a vector representation in \(\mathbb{R}^d\).
    \item \( e: \text{Word} \to \mathbb{R}^b \) maps a word to a vector representation in \(\mathbb{R}^b\).
\end{itemize}

Our model is a neural network \( m: \mathbb{R}^d \to \mathbb{R}^b \) optimized through extensive experimentation, which learns a transformation between the definition and word embeddings:
\begin{equation}
    m(f(\text{def}_i)) \approx e(\text{word}_i).
\end{equation}

The architecture follows a semi-encoder structure with four hidden layers, where the layer sizes decrease geometrically by a factor of 2, starting from \( 8s \) (where \( s = d = 256 \)) down to \( s \):
\[
h_1 = 8s, \quad h_2 = 4s, \quad h_3 = 2s, \quad h_4 = s.
\]

The model is trained using the Mean Squared Error (MSE) loss:
\begin{equation}\label{eq_MSE}
    \mathcal{L} = \frac{1}{|D|} \sum_{i \in I} \| m(f(\text{def}_i)) - e(\text{word}_i) \|^2.
\end{equation}

We employ the GELU activation function for non-linearity and apply an adjustable dropout rate \( d \in [0.2, 0.4] \) to mitigate overfitting. The AdamW optimizer is used with a learning rate of \(\eta = 1 \times 10^{-4} \), as higher values (e.g., \( 1 \times 10^{-3} \)) led to instability in previous experiments and this architicture is chosen after empirical expirements from \cite{sibaee-etal-2023-qamosy}.

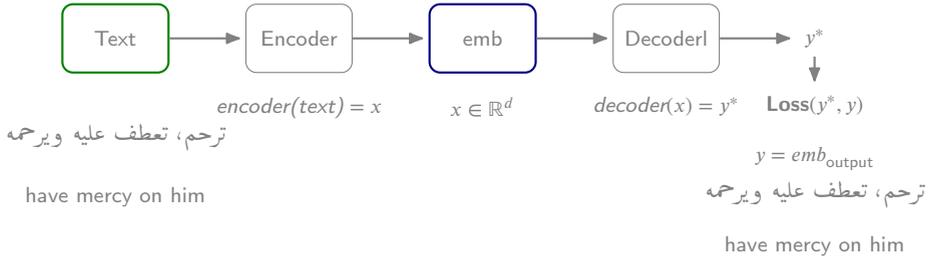
\begin{figure}[ht]
\centering
\begin{tikzpicture}[
    node distance=1.2cm and 1.0cm,
    box/.style={draw, rounded corners, minimum width=1.4cm, minimum height=0.9cm, align=center},
    textnode/.style={draw=green!50!black, thick, rounded corners, minimum width=1.4cm, minimum height=0.9cm},
    embnode/.style={draw=blue!50!black, thick, rounded corners, minimum width=1.4cm, minimum height=0.9cm},
    arrow/.style={-{Latex[round]}, thick, shorten >=2pt},
    every node/.append style={font=\footnotesize}
]

% Nodes
\node[textnode] (text) {Text};
\node[box, right=of text] (encoder) {Encoder};
\node[embnode, right=of encoder] (emb) {emb};
\node[box, right=of emb] (model) {Decoderl};
\node[right=of model] (ystar) {$y^*$};
\node[below=0.4cm of ystar] (loss) {$\textbf{Loss}(y^*, y)$};

% Arrows
\draw[arrow] (text) -- (encoder);
\draw[arrow] (encoder) -- (emb);
\draw[arrow] (emb) -- (model);
\draw[arrow] (model) -- (ystar);
\draw[arrow] (ystar) -- (loss);

% Equations under each node
\node[below=0.5cm of text] (lbl1) {\<ترحم، تعطف عليه ويرحمه>};
\node[below=0.2cm of lbl1] {have mercy on him};

\node[below=0.2cm of encoder] {$\textit{encoder(text)} = x$};
\node[below=0.2cm of emb] {$x \in \mathbb{R}^d$};
\node[below=0.2cm of model] {$\textit{decoder}(x) = y^*$};
\node[below=0.2cm of loss] {$y = emb_{\text{output}}$};

\node[below=0.6cm of loss] (lbl2) {\<ترحم، تعطف عليه ويرحمه>};
\node[below=0.1cm of lbl2] {have mercy on him};

\end{tikzpicture}
\caption{Overview of the encoder-decoder flow with embedding and target loss.}
\label{fig:process_sum}
\end{figure}

The described architecture is consistent across all experiments, with modifications restricted to dropout rates and optimizer settings. The compact training process is shown in Figure ~\ref{fig:process_sum}.

% \begin{figure}[!h]
%     \centering
%     \includegraphics[width=0.97\textwidth]{Reseach-RD graph.drawio.pdf}
%     \caption{Process of the Training}
%     \label{fig:process_sum}
% \end{figure}

\subsubsection{Inference and Ranking-Based Evaluation}

Given a test definition \( \text{def}' \), we retrieve the most relevant word by ranking candidate words based on a similarity metric.

\begin{itemize}
    \item \textbf{Similarity Computation}

    We define a similarity function \( S: \mathbb{R}^b \times \mathbb{R}^b \to \mathbb{R} \), such as cosine similarity:
    \begin{equation}
        S(x, y) = \frac{x \cdot y}{\|x\| \|y\|}
    \end{equation}

    For each candidate word \( w \in V \), we compute:
    \begin{equation}
        S_w = S(m(f(\text{def}')), e(w)).
    \end{equation}

    \item \textbf{Rank Assignment}

    For each word \( w \), a set of metric scores \( M_w = \{M_1(w), M_2(w), \dots, M_k(w)\} \) is computed, where each \( M_i(w) \) represents a different evaluation metric (e.g., MSE, cosine similarity, rank).

    The rank transformation is defined as:
    \begin{equation}
        R_w = \sum_{x \in M_w} \mathsf{1} (x \geq M_w) \quad \text{if the metric is to be maximized,}
    \end{equation}
    \begin{equation}
        R_w = \sum_{x \in M_w} \mathsf{1} (x \leq M_w) \quad \text{if the metric is to be minimized.}
    \end{equation}
    where \( \mathsf{1}(\cdot) \) is the indicator function.

    \item \textbf{Final Ranking Score}

    For each language \( \ell \), the final ranking score is computed as the mean of all individual ranks $k$:
    \begin{equation}
        \text{Final Rank}_\ell = \frac{1}{k} \sum_{i=1}^{k} R_{i,w}.
    \end{equation}

    The final rankings are aggregated across users and stored for comparison.
\end{itemize}

\section{Results}
In this section we show our results (trained on the model shown in Section 3).
% \textcolor{red}{This section may be divided by subheadings. It should provide a concise and precise description of the experimental results, their interpretation as well as the experimental conclusions that can be drawn.}

\subsection{Models results}
In this section we show the results of all tested models in terms of MSE, cosine similarity, and rank (Table~\ref{tab:model_results}).
\begin{table}[!h]
    \centering
    \begin{tabular}{lccc}
        \toprule
        \textbf{Model} & \textbf{MSE} & \textbf{Cosine Similarity} & \textbf{Rank} \\
        \midrule
        baselineM\_elct \cite{alshammari-etal-2024-ksaa}  & \textbf{0.145} & \textbf{0.736} & 0.84 \\
        MARBERTv2 \cite{abdul-mageed-etal-2021-arbert} & 0.227 & 0.559 & 0.174 \\
        ARBERTv2 \cite{abdul-mageed-etal-2021-arbert} & 0.158 & 0.7071 & \textbf{0.0644} \\
        OpenAI large  & 0.177 & 0.673 & 0.077 \\
        OpenAI small & 0.187 & 0.649 & 0.097 \\
        Gate v1 \cite{nacar2025GATE} & 0.208 & 0.596 & 0.221 \\
        ATMv2 \cite{nacar2024enhancingsemanticsimilarityunderstanding} & 0.222 & 0.570 & 0.183 \\
        Nomic-embed-text-v2 \cite{nussbaum2025trainingsparsemixtureexperts} & 0.215 & 0.587 & 0.145 \\
        mpnet \cite{reimers2020makingmonolingualsentenceembeddings} & 0.225 & 0.562 & 0.22 \\
        L12 \cite{reimers2020makingmonolingualsentenceembeddings} & 0.220 & 0.560 & 0.22 \\
        LaBSE \cite{reimers2020makingmonolingualsentenceembeddings} & 0.221 & 0.570 & 0.17 \\
        CasedV1 \cite{reimers2020makingmonolingualsentenceembeddings} & 0.226 & 0.560 & 0.21 \\
        Snowflake-arctic-embed \cite{yu2024arcticembed20multilingualretrieval} & 0.207 & 0.607 & 0.117 \\
        \bottomrule
    \end{tabular}
    \caption{Comparison of model performance based on MSE, cosine similarity, and rank.}
    \label{tab:model_results}
\end{table}

\section{Discussion}
In this section we anaylze and discuss the results and models performences from section 4.
\subsection{Analyzing models performance}
The extended experiments further validate our initial findings regarding Arabic language models. ARBERTv2 achieves the best rank metric (0.0644), which is the primary evaluation criterion for our study. This superior rank performance demonstrates ARBERTv2's exceptional ability to preserve relative semantic distances between embeddings in Arabic text. While baselineM\_elct shows the lowest MSE (0.145) and highest cosine similarity (0.736), ARBERTv2's rank performance indicates its particular strength in maintaining semantic relationships that are crucial for downstream applications.
The rank metric results reveal a significant performance gap between Arabic-specific models and general multilingual embeddings. OpenAI models (large and small) perform competitively with rank scores of 0.077 and 0.097 respectively, suggesting that large-scale training across multiple languages can partially compensate for language-specific optimization. However, their rank scores still trail behind ARBERTv2, confirming the advantage of language-specialized training.
General embedding models (Nomic-embed-text-v2, Matyroshka, mpnet) show substantially weaker rank performance (0.145-0.22), further emphasizing the importance of language-specific training for optimal semantic representation. MARBERTv2's poor rank performance (0.174) compared to ARBERTv2 highlights that architectural variations within the same model family can significantly impact embedding quality.
These findings reinforce our conclusion that the choice of pre-training data and optimization strategy is crucial for developing effective language-specific embedding models, particularly when evaluated on the rank metric that best captures semantic relationship preservation.

\subsection{Analyzing the Dataset}

In this section, we outline the key insights derived from analyzing more than 200 randomly selected samples from the dataset (with an average score \( \frac{3}{5}\) as depicted in Figure \ref{fig:count_acc}). Each sample was evaluated on a scale of 1 to 5, leading to the following observations:

\begin{figure}[!h] % 'p' forces it to a separate page
    \centering
    \includegraphics[width=0.7\textwidth]{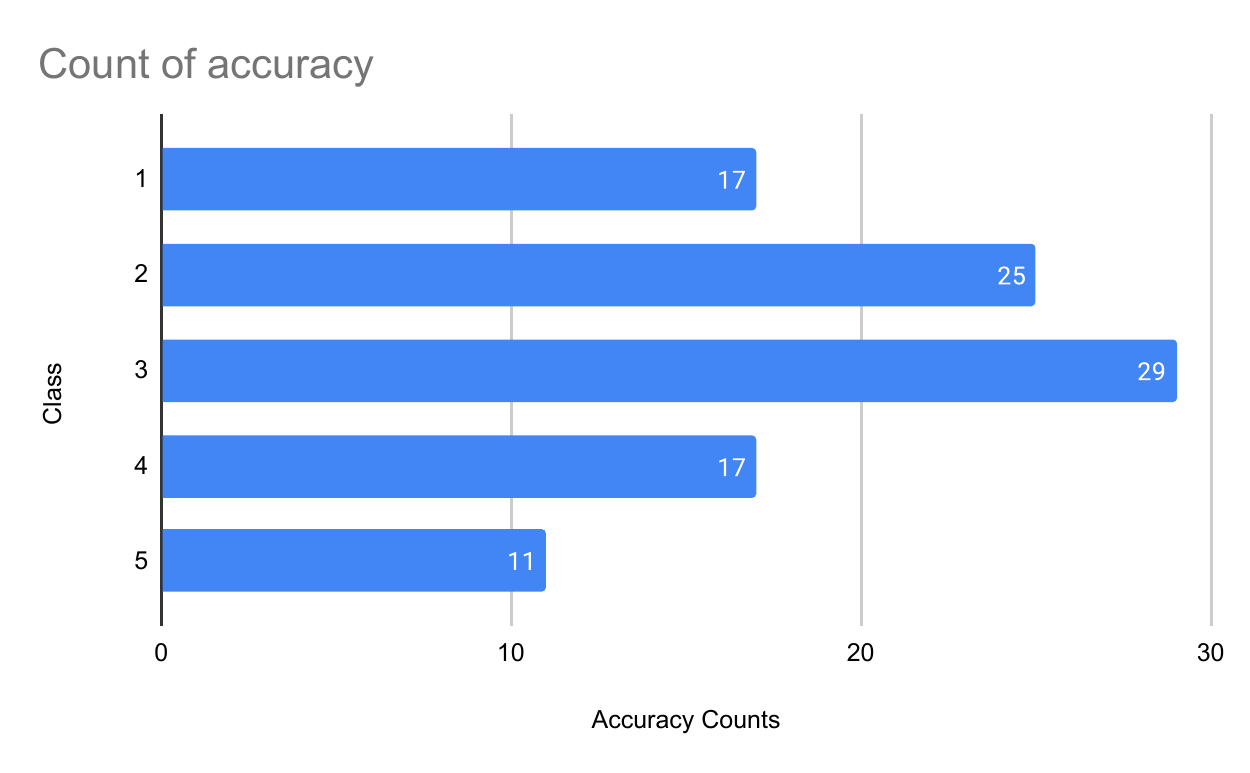} % Adjust width (smaller)
    \caption{Accuracy distribution}
    \label{fig:count_acc}
\end{figure}

Lack of a systematic and logical methodology for definitions. The definitions exhibit significant variability in formulation. Some focus on morphological forms, while others emphasize a specific meaning derived from the general meaning of the Arabic root. Additionally, some definitions are extracted from specialized scientific contexts without specifying the targeted field. Furthermore, certain definitions include the word(s) intended to be defined within the definition itself, which is an unclear approach. Examples illustrating these issues are detailed below.

\begin{enumerate}
    \item Reliance on morphological forms rather than defining meanings.
    
    Several definitions rely on morphological derivations rather than providing clear meanings. For instance, the word \<تشمل> (to include) is defined by listing its morphological derivatives: \<تشمّلاً، فهو مُتشمّل والمفعول مُتشمّل به> (inclusion, one who includes, and the object being included), none of which convey the actual meaning. Similar cases include \<ينوع> (to vary), \<صعد> (to ascend), and \<تحالف> (to form an alliance), where only verbal nouns and inflections are mentioned instead of semantic explanations. Our evaluation indicates that more than 30\% of the dataset follows this pattern.
    
    \item Use of pronouns without clear reference to the defined word.
    
    Some definitions employ pronouns without clearly linking them to the term being defined. For example, in \<ضعفها قُصور الذّاكرة> (its weakness is memory deficiency), the pronoun \<ها> (its) is ambiguous, making the definition unclear.
    
    \item Overly specific definitions that do not reflect general meanings.
    
    Some definitions are excessively specialized, restricting the broader meaning of a term. For example:
    \begin{itemize}
        \item \<مِنْقاش> (tweezers) is defined as \<أداة يستخرج بها الرّاصف الحروفَ> (a tool used by a typesetter to extract letters), which is a rare technical definition, whereas its more common meaning is \<أداة لإزالة الشوائب من الجلد> (a tool for removing impurities from the skin).
        \item \<شكليّة> (formalism) is defined within a legal context as \<مبدأ قوامه أنّ صحّة الأعمال القانونيّة> (a principle stating that the validity of legal acts...), whereas it is more commonly used in philosophy and the arts.
        \item \<حقّق مع فلان> (to interrogate someone) is narrowly defined in legal contexts as \<أخذ أقوالَه في قضيّةٍ ما> (taking someone's statements in a case), though \<حقّق> (to investigate) is also widely used in manuscript studies and literary contexts.
    \end{itemize}
    
    \item Definitions that are too field-specific without indicating general meaning.
    
    Many definitions are provided from specialized disciplinary perspectives without considering broader meanings. For example:
    \begin{itemize}
        \item \<عرفي> (customary) is given only its legal definition, ignoring its general usage.
        \item \<بدن> (body) is defined as \<بدن الثوب> (the main part of a garment), which is a narrow textile-related meaning instead of the more common \<الجسم> (body).
        \item \<وتر> (chord) is defined in a mathematical sense as \<وتر المثلث> (the hypotenuse of a triangle) rather than the more general meaning \<وتر القوس حبله> (the bowstring of a bow).
    \end{itemize}
    
    \item Definitions using illustrative phrases instead of direct explanations.
    
    Some definitions rely on idiomatic expressions rather than clearly defining the word itself. For example:
    \begin{itemize}
        \item \<رقبتي سَدَّادة> (my neck is a stopper) is given instead of a direct definition of the intended meaning, the defnition should include the information of this idiomatic expression then explaining it. 
    \end{itemize}
    
    \item Use of redundant phrasing in defined terms.
    
    Some terms (defined words) use unnecessarily long formulations when a shorter one would suffice. Examples include:
    \begin{itemize}
        \item \< تنفَل المُصلِّي > (a worshipper performs extra prayers) is defined as
        
        whereas simply stating \<تنفل> (performed extra prayers) would be clearer.
        
        \item \<رست المناقصةُ على فلان> (the bid was awarded to someone) is defined as
        
        Though \<رست المناقصة> (the bid was awarded) alone is sufficient and provides adequate context.
    \end{itemize}
    
    \item Use of synonyms instead of actual definitions.
    
    Some definitions list synonyms rather than providing a precise explanation. For example:
    \begin{itemize}
        \item \<مُهْمِل،‏ ومتقاعس> (negligent and sluggish) is given as the definition of \<مُتَهَاوِن> (careless), which does not clarify its meaning.
    \end{itemize}
\end{enumerate}

\subsubsection{Standards for Building RD Dataset}

\begin{figure}[!h] % 'p' forces it to a separate page
    \centering
    \includegraphics[width=0.5\textwidth]{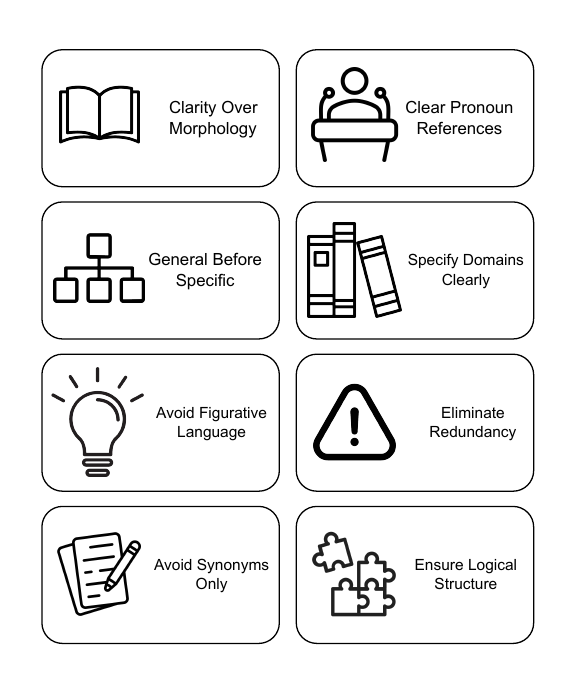} % Adjust width (smaller)
    \caption{Summary of RD writing standards.}
    \label{fig:std_sum}
\end{figure}

From the lessons learned from the previous experiments, and to ensure high-quality dataset construction for a reverse dictionary application in Arabic, the following standards (which was extracted after analyzing random samples from the dataset) should be maintained. Standards are summarized in Figure \ref{fig:std_sum}:

\begin{enumerate}
    \item \textbf{Definitions should provide clear meanings rather than morphological forms.} Avoid listing derivatives and inflections without conveying the actual meaning of the term. Instead, focus on explicit semantic explanations. 
    
    \item \textbf{Pronoun references should be unambiguous.} Ensure that pronouns have a clear and direct connection to the defined word to prevent ambiguity in interpretation.
    
    \item \textbf{Definitions should reflect general meanings before specialized ones.} A definition should first cover the broadest possible meaning before specifying context-dependent interpretations, ensuring wider applicability.
    
    \item \textbf{Indicate specialized meanings explicitly.} When providing a definition from a specific field (e.g., legal, mathematical, or medical contexts), explicitly state the domain to avoid confusion.
    
    \item \textbf{Avoid using illustrative phrases instead of definitions.} Definitions should be direct and explanatory rather than relying on idiomatic expressions or metaphors.
    
    \item \textbf{Eliminate redundant phrasing.} Use concise and precise language to convey meaning efficiently, avoiding unnecessary elaboration.
    
    \item \textbf{Use accurate and informative descriptions instead of synonyms.} A definition should clarify the meaning rather than merely listing synonyms, ensuring that users fully understand the concept.
    
    \item \textbf{Maintain logical consistency and systematic structuring.} Definitions should follow a uniform and logical methodology, ensuring consistency throughout the dataset.
\end{enumerate}

\section{Conclusions}
This study presented the development of an Arabic Reverse Dictionary (RD) system that addresses significant challenges in Arabic natural language processing. Our semi-encoder neural network architecture with geometrically decreasing layers achieved state-of-the-art results, demonstrating the effectiveness of language-specialized models for semantic representation tasks. The comparative analysis of various embedding models confirms that Arabic-specific pre-trained models, particularly ARBERTv2, significantly outperform general multilingual embeddings in maintaining semantic relationships essential for RD applications.
Our detailed analysis of the dataset revealed critical insights into the quality and consistency of Arabic lexicographic definitions, leading to the establishment of eight formal standards for constructing high-quality RD resources. These standards emphasize clarity over morphology, unambiguous references, general before specific meanings, domain specification, avoidance of figurative language, elimination of redundancy, comprehensive descriptions over mere synonyms, and logical structure.
The modular RDTL library we developed provides researchers and developers with configurable training pipelines for RD tasks, facilitating further innovation in this field. Future work should focus on implementing these definition standards in larger datasets, exploring multi-task learning approaches that combine RD with related semantic tasks, and investigating methods to handle Arabic's rich morphological structures more effectively.
By advancing Arabic RD systems, this work contributes to broader accessibility of precise linguistic tools for Arabic speakers, supporting applications in education, academic writing, professional communication, and general language use. The methods and insights presented here lay a foundation for continued progress in Arabic computational linguistics and semantic understanding technologies.

\section{Acknowledgment}
The authors would like to acknowledge the support of Prince Sultan University for paying the Article
Processing Charges (APC) of this publication.

\printcredits

%% Loading bibliography style and database
\bibliographystyle{cas-model2-names}
\bibliography{main}

\end{document}